\documentclass[10pt,journal,compsoc]{IEEEtran}

\usepackage{hyperref}       
\usepackage{url}            
\usepackage{booktabs}       
\usepackage{amsfonts}       
\usepackage{nicefrac}       
\usepackage{microtype}      
\usepackage{enumerate}
\usepackage{graphicx}
\usepackage{subfigure}
\usepackage{booktabs}
\usepackage{multirow}
\usepackage{amsmath}
\usepackage{amssymb}
\usepackage{graphicx}
\usepackage{bbm}
\usepackage{subfigure}
\usepackage{wrapfig}
\usepackage{hyperref}
\usepackage{pifont}
\usepackage{color}
\usepackage[ruled]{algorithm2e}
\usepackage{algorithmic}
\usepackage{amsmath}
\usepackage{amsthm}
\usepackage[switch]{lineno}
\usepackage{bm}

\newtheorem{theorem}{Theorem}

\newcommand{\xmark}{\ding{55}}
\definecolor{mycolor2}{rgb}{0.0,0.51,0.22}
\definecolor{deepred}{rgb}{0.64, 0.0, 0.0}
\definecolor{customblue}{rgb}{0.18, 0.46, 0.71}
\newcommand{\improved}[2]{$#1\%$ \textcolor{mycolor2}{$\Uparrow_{#2\%}$}}
\newcommand{\relimproved}[3]{$#1.#2\%$ \textcolor{mycolor2}{$\Uparrow_{#3\%}$}}
\usepackage[most]{tcolorbox}

\usepackage{setspace}

\usepackage{ragged2e}
\usepackage{floatrow}
\newfloatcommand{capbtabbox}{table}[][\FBwidth]
\usepackage{tikz}

\usepackage{array}
\newcolumntype{L}[1]{>{\raggedright\let\newline\\\arraybackslash\hspace{0pt}}m{#1}}
\usepackage{paralist}
\let\itemize\compactitem
\let\enditemize\endcompactitem
\let\enumerate\compactenum
\let\endenumerate\endcompactenum

\ifCLASSOPTIONcompsoc
  \usepackage[nocompress]{cite}
\else
  \usepackage{cite}
\fi

\ifCLASSINFOpdf
\else
\fi

\begin{document}
%

\title{Estimating LLM Uncertainty with Evidence}

\author{{Huan Ma$^1$, Jingdong Chen$^1$, Joey Tianyi Zhou$^3$, Guangyu Wang$^2$, Changqing Zhang$^1$ }
\IEEEcompsocitemizethanks{\IEEEcompsocthanksitem $^1$Tianjin University (zhangchangqing@tju.edu.cn), $^2$Beijing University of Posts and Telecommunications (guangyu.wang24@gmail.com), $^3$A*STAR, Singapore.}
}

%
%

\markboth{Estimating LLM Uncertainty with Logits}%
{}

\IEEEtitleabstractindextext{%
\begin{abstract}
Over the past few years, Large Language Models (LLMs) have developed rapidly and are widely applied in various domains. However, LLMs face the issue of hallucinations, generating responses that may be unreliable when the models lack relevant knowledge. To be aware of potential hallucinations, uncertainty estimation methods have been introduced, and most of them have confirmed that reliability lies in critical tokens. However, probability-based methods perform poorly in identifying token reliability, limiting their practical utility. In this paper, we reveal that the probability-based method fails to estimate token reliability due to the loss of evidence strength information which is accumulated in the training stage. Therefore, we present Logits-induced token uncertainty (LogTokU), a framework for estimating decoupled token uncertainty in LLMs, enabling real-time uncertainty estimation without requiring multiple sampling processes. We employ evidence modeling to implement LogTokU and use the estimated uncertainty to guide downstream tasks. The experimental results demonstrate that LogTokU has significant effectiveness and promise. Our code is available at \href{https://github.com/MaHuanAAA/logtoku}{link}.

\end{abstract}

}

\maketitle

\IEEEdisplaynontitleabstractindextext

%
\IEEEpeerreviewmaketitle

\section{Introduction}
Over the past few years, Large Language Models (LLMs) have developed rapidly and LLM-driven systems are deployed across various domains. Despite their remarkable performance, LLMs remain prone to hallucinations~\cite{banerjee2024llms}, which causes them to generate unreliable responses when the models lack the corresponding knowledge. Hallucinations critically undermine the reliability of LLMs, particularly in professional applications such as medical and legal consultations~\cite{shah2024accuracy,dahl2024large}. Hallucinations have been considered as a major barrier to the broader deployment of LLMs~\cite{huang2023survey,liu2024exploring,perkovic2024hallucinations,zhou2024larger}.

\begin{figure*}
\centering 
\subfigure[Current probability-based methods fail in estimating response reliability.]{ \includegraphics[width=0.61\textwidth]{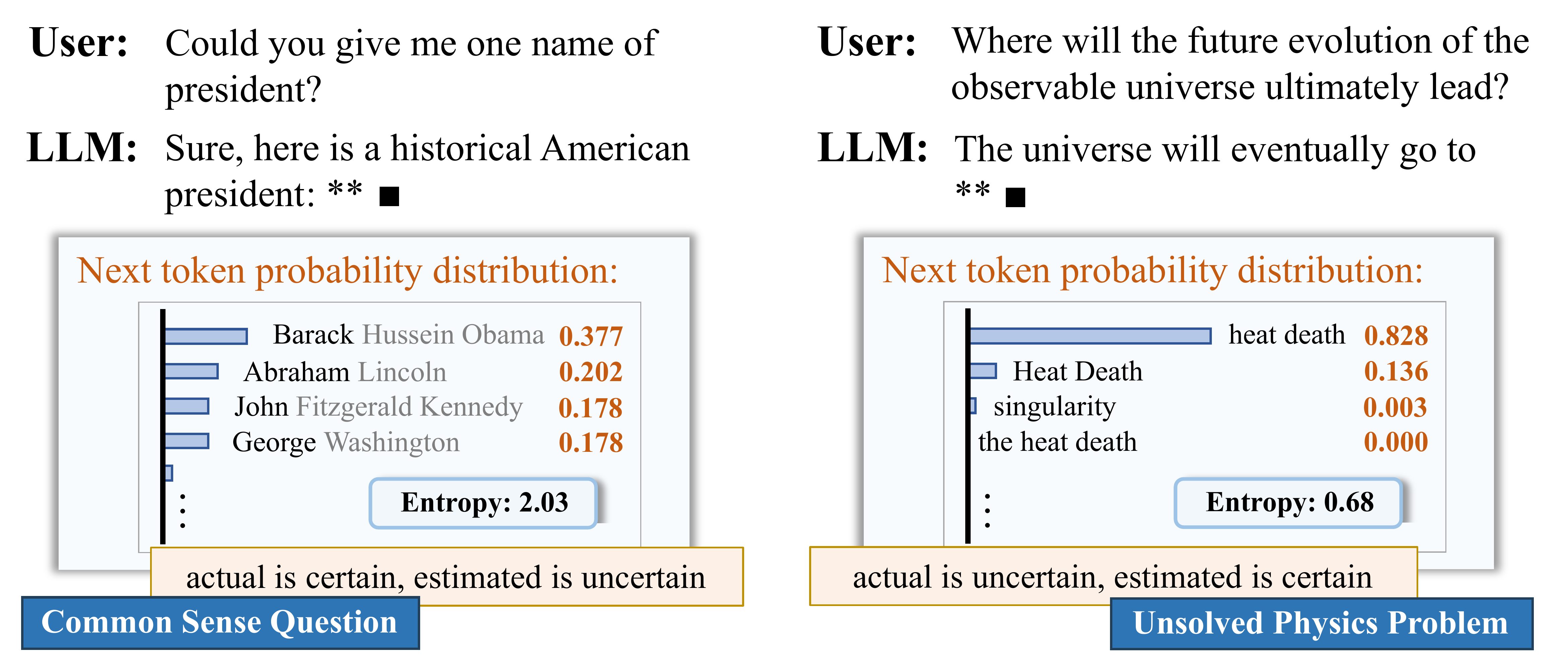} \label{fig:probability}
} \hfill 
\subfigure[Comparison of logits and probability.]{ \includegraphics[width=0.32\textwidth]{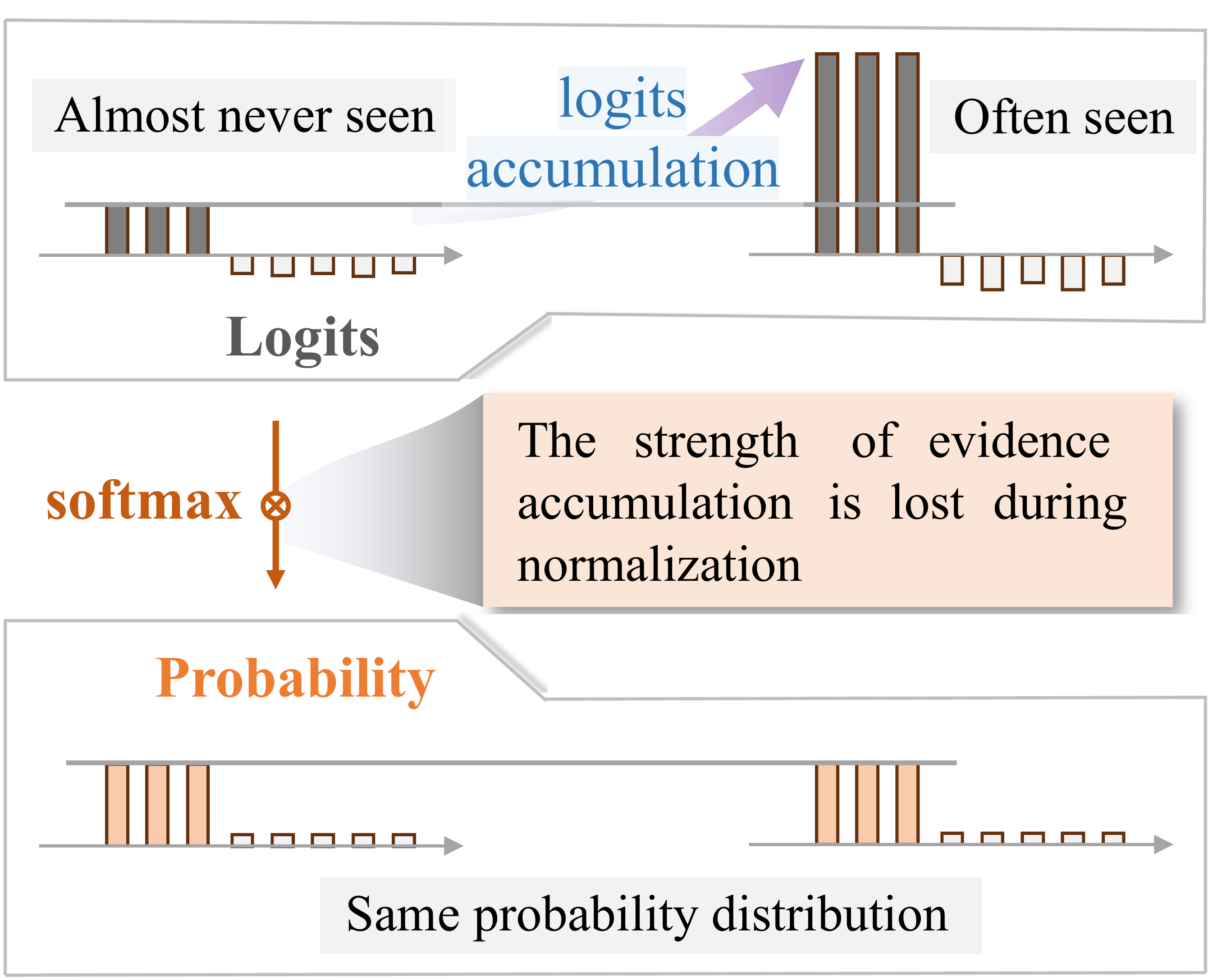} 
\label{fig:softmax} }
\caption{\textbf{Why probability-based methods fail?} \emph{Left}: A pair of examples on LLaMA-2 shows that probability fails in estimating reliability. Since LLMs know the names of many presidents, the probability after normalization is very low; whereas for the future of the universe, since LLMs only know one hypothesis, the probability is very high. The probability-based reliability measure is counterintuitive, as the answers on common sense questions, where LLMs have rich knowledge, are less reliable than on unsolved physics problems. This is because probability cannot reflect whether a low probability is due to LLMs knowing multiple correct answers. These two cases are well characterized in this paper, corresponding to the fourth and second quadrants of Fig.~\ref{fig:cover1}, respectively. \emph{Right}: Normalization leads to the loss of evidence strength information.} 
\end{figure*}

Uncertainty estimation has shown promise in the identification of hallucinations in LLMs~\cite{xiao2021hallucination,huang2024survey}. High uncertainty often indicates the need for caution from users, as it suggests that the model is likely to be influenced by hallucinations~\cite{zhang2023enhancing,yoffe2024debunc}. In other words, high uncertainty signals unreliable responses. However, existing methods for uncertainty estimation in LLMs have limitations in estimating the inherent uncertainty of LLMs and lack efficiency. Including discussions of various heuristic methods, such as self-reflection~\cite{ji2023towards}, LLM uncertainty estimation can be categorized into sampling-based and token-based methods. Sampling-based methods estimate uncertainty by multiple sampling~\cite{liu2024exploring}, perturbations~\cite{zhang2024sac3reliable}, or contrastive analysis~\cite{huang2024uncttp}. Semantic Entropy (SE)~\cite{kuhn2023semantic} is a representative technique among these sampling-based methods. The core of these methods is estimating the consistency of LLM's multiple guesses, and the response will be marked as unreliable if inconsistency is detected. However, sampling-based methods suffer from several limitations: (1) it cannot assess the reliability of a single response; (2) it requires multiple sampling iterations, making it inefficient and impractical for deployment in real-world applications; and (3) it fails to account for the model's inherent uncertainty, such as consistently incorrect responses due to a lack of knowledge. 

Token-based methods can estimate the uncertainty of a single sentence without requiring multiple samplings as deterministic approaches~\cite{gupta2024language,fadeeva2024fact}. However, due to the lack of effective token-level uncertainty estimation, these methods often fail to achieve satisfactory results. For example, many tokens are correct but exhibit low probabilities (high entropy) in the model response.
Many works have pointed out that the importance of different tokens in a sentence is not uniform~\cite{lin2024critical,duan2024shifting}, and the reliability of the model response depends mainly on a few critical tokens~\cite{duan2024gtbench,bigelow2024forking}, so they only focus on critical tokens when estimating reliability. However, as shown in Fig.~\ref{fig:probability}, although both ``\texttt{Barack Hussein Obama}'' and ``\texttt{Abraham Lincoln}'' could be correct answers, their maximum probability of the critical token (first token of their names) is only $0.377$, which indicates a very high uncertainty. Therefore, to obtain an LLM uncertainty estimation that is both effective and efficient, a reliable token-level uncertainty estimation method is urgently required.

\begin{figure*}
\centering 
\subfigure[Logits-induced token uncertainty (LogTokU).]{ \includegraphics[width=0.43\textwidth]{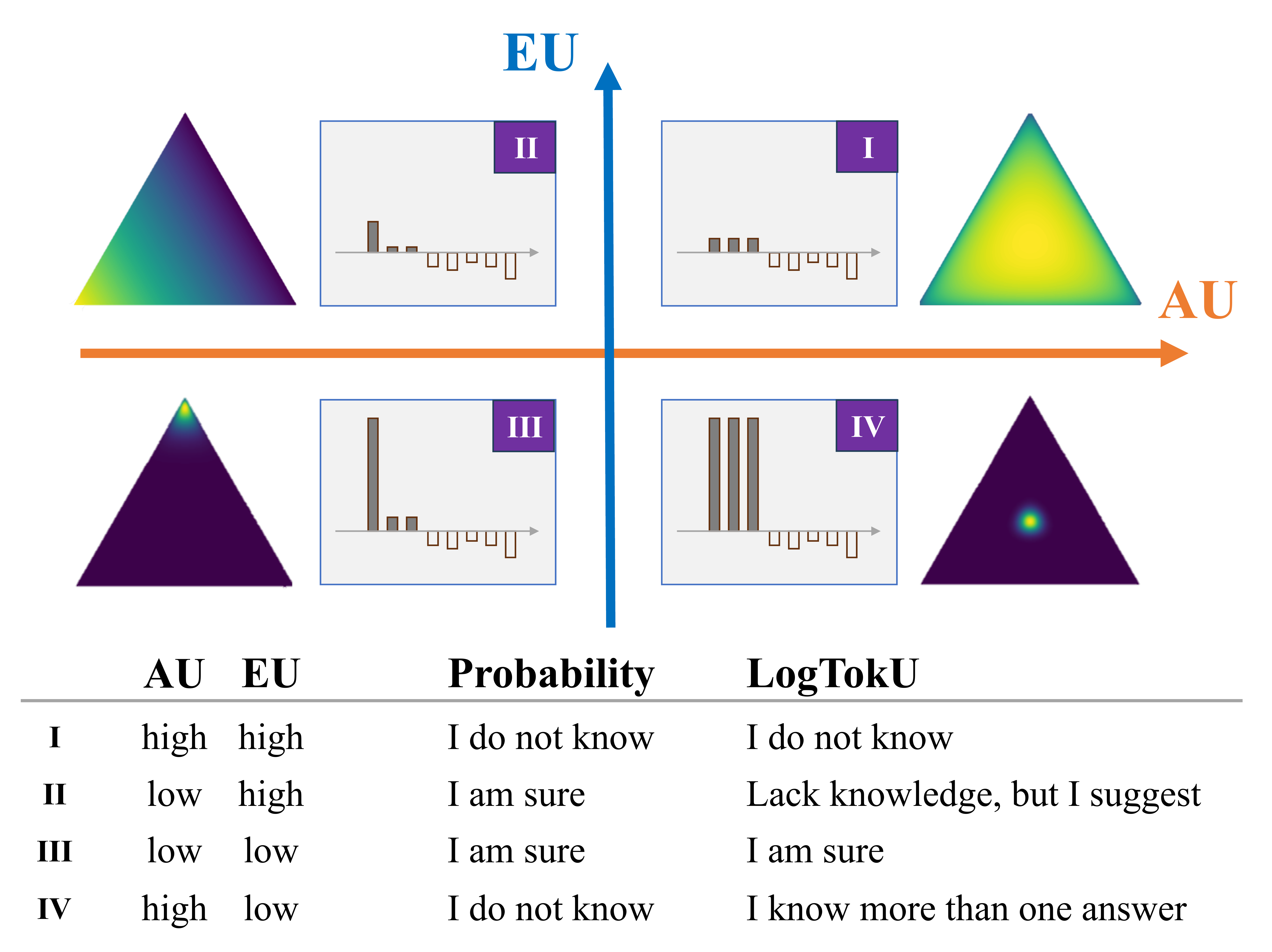} \label{fig:cover1}
} \hfill 
\subfigure[An example dialogue on LLaMA-2-Chat.]{ \includegraphics[width=0.51\textwidth]{figs/case.pdf} 
\label{fig:cover2} }
\caption{\textbf{Why LogTokU works?} \emph{Left}: Illustration of four different scenarios considered in LogTokU, where the gray bars represent the logits for predicting the next token, the triangular patterns represent the corresponding Dirichlet distribution, and the table below compares uncertainty estimation using probability with that using LogTokU. \emph{Right}: A case study from a medical QA, where the markings under each word reflect reliability estimated according to LogTokU, as well as the values of AU (gray) and EU (blue). \textbf{I}: Both AU and EU are high, where LLaMA recommends a metal ``\texttt{Chromium}'' for diabetes patients. \textbf{II}: The total logits are low, but one token's logit is larger than the others, indicating that the LLM lacks experience and knowledge but knows what should be the next token, where the LLM repeats the medicine ``\texttt{Glucomannan}'' that has been generated in the previous context. \textbf{III}: The LLM is very confident about the next token, where it generates the fixed phrase ``has \texttt{been}''. \textbf{IV}: The LLM has enough knowledge and knows more than one suitable answer. For example, the LLM generates ``\texttt{[comma]}'', which can be replaced by many other suitable words. The dilemma in Fig. 1 is addressed according to quadrant II and IV.} \label{fig:two_subfigures}
\end{figure*}

In this paper, we find that the reason why the current token uncertainty estimation fails is that probability no longer captures the reliability of LLM responses. Specifically, probability can only reflect the relative strength relationship between different categories from the perspective of a discriminative model. However, generative LLMs are different from traditional discriminative models, and there may be more than one token that can be the suitable next-token. Therefore, measuring the probability of a single suitable candidate cannot capture the reliability of the model response. To capture the reliability of generating the next token, we need to assess whether the model has the knowledge to generate the next token, which can be referred to as the model's inherent uncertainty. In this paper, we reveal that the strength of evidence accumulation is lost during normalization, leading to the failure of probability-based reliability estimation. Therefore, we present a new uncertainty estimation framework based on logits named \textbf{Log}its-induced \textbf{Tok}en \textbf{U}ncertainty (LogTokU), which achieves both effectiveness and efficiency. Specifically, LogTokU introduces a more clear token-level uncertainty estimation, decoupling uncertainty into tokens relative aleatoric uncertainty (AU) and model inherent epistemic uncertainty (EU). As illustrated in Fig.~\ref{fig:cover1}, distinct combinations of these two types of uncertainty are treated as separate cases. Compared to probability-based uncertainty modeling, LogTokU can not only express states of ``I am sure'' and ``I do not know'', but it can also express ``Lack knowledge, but I suggest'' and ``I know more than one answer''. Moreover, this framework enables the estimation of real-time reliability of any response without the need for sampling, ensuring high efficiency. To realize the LogTokU framework, we adapt evidence modeling~\cite{sensoy2018evidential} in this paper, treating logits as parameters of a Dirichlet distribution to characterize aleatoric uncertainty and epistemic uncertainty. Then, we validate the reliability of the estimated uncertainty by using it to guide downstream tasks.
The main contributions of this paper can be summarized as: (1) We reveal why probability-based strategies fail and introduce \textbf{a new framework} to estimate the token uncertainty of LLMs, which shows that the token uncertainty is a promising way to estimate response reliability. (2) By employing evidential learning to estimate aleatoric uncertainty and epistemic uncertainty, we provide \textbf{a viable implementation} of the LogTokU framework. 
(3) We demonstrate \textbf{two usage cases} for the estimated uncertainty, thus validating its effectiveness and showcasing the significant potential of LogTokU.

\section{Related Work}

\textbf{Sampling-based uncertainty estimation}.\quad  
These methods evaluate the randomness in the LLM generation process. Specifically, they allow LLMs to guess multiple times and evaluate their consistency. \cite{selectivelyanswering2023} introduce repetition and diversity into the measurement of consistency.
\cite{zhang2024sac3reliable} introduce a mechanism that perturbs semantically equivalent questions to evaluate the consistency of LLM responses across variants of the same question.  
\cite{huang2024uncttp} inject correct and incorrect labels into the prompt during sampling, and the uncertainty level is defined based on whether the LLM's responses remain consistent across three samplings for each instance.  
A more continuous measurement of uncertainty is based on similarity. A recent method proposed by \cite{lin2024generating} calculates the similarity between multiple responses to indirectly quantify the dispersion of the model outputs.  
The most representative sampling-based method is Semantic Entropy (SE)~\cite{kuhn2023semantic}, which improves token-level measures by clustering sentences into equivalence classes based on their semantic similarity and computing the entropy over these classes. However, sampling-based methods can not evaluate the model's inherent uncertainty and are costly. Specifically, sampling-based methods only measure the consistency of multiple guesses but fail to identify whether the LLM lacks knowledge about the question.

\textbf{Probability-based uncertainty estimation}.\quad  
Probability-based uncertainty estimation methods can also be described as deterministic methods. These methods calculate the confidence (uncertainty) of the model based on the probability distribution of the prediction~\cite{arora2021prob2}. In addition to maximum probability, entropy is another common measure for estimating uncertainty. \cite{kadavath2022language} use the probability of the complete sequence to compute predictive entropy for assessing the sharpness of the output distributions. However, not all tokens in a sentence are equally critical~\cite{lin2024critical}. Recent work by \cite{duan2024shifting} highlights that not all tokens contribute equally to the underlying meaning, as linguistic redundancy often allows a few key tokens to capture the essence of longer sentences. However, even when focusing only on key tokens, these probability-based methods still cannot estimate the reliability of the answer generated by LLM. As shown in Fig.~\ref{fig:probability}, the key tokens do not yet provide a reliable estimation of the risk.

\section{Logits-induced Token Uncertainty}

\subsection{Notations}
Consider that given a pre-trained LLM noted as $\mathcal{M}$ and its corresponding tokenizer dictionary $\bm{Y}=\{\tau^1,\tau^2,\cdots,\tau^{|\bm{Y}|}\}$, and $|\bm{Y}|$ indicates the size of the vocabulary dictionary. Specifically, the user inputs an instruction, then the instruction is transformed into a prompt by applying a chat template (for example, ``\texttt{[INST]Could you give me one name of president?[\textbackslash INST]}''. The prompt is encoded by the corresponding tokenizer as a vector $\bm{q}$ and input into LLM to perform the next token prediction under certain basic sampling strategies. The model continuously generates the next token $a_{t}$ based on the $\bm{q}$ and tokens that have been generated $\bm{a}_{t-1}=a_{1}a_{2}\cdots a_{t-1}$ (for example, $\bm{a}_{t-1}$ is a generated token vector can be decoded into ``\texttt{Sure, here is a historical American president:**}'') until they meet the stop rule (for example, meeting \texttt{[EOS]}), which can be formulated as:
\begin{equation}
\begin{aligned}
    \textbf{P}(\bm{Y}|\bm{q},&\bm{a}_{t-1},\mathcal{M}) = \{ p(\tau^{m}|\bm{q},\bm{a}_{t-1},\mathcal{M})\}_{m=1}^{|Y|}\\=&\left\{\frac{\exp(\mathcal{M}(\tau^{m}|\bm{q},\bm{a}_{t-1}))}{\sum_{j=1}^{|\bm{Y}|}\exp(\mathcal{M}(\tau^{j}|\bm{q},\bm{a}_{t-1}))}\right\}_{m=1}^{|\bm{Y}|},      
\end{aligned}
\end{equation}
where $\mathcal{M}(\tau^{m}|\bm{q},\bm{a}_{t-1})$ indicates the predicted logit (score before softmax layer) of $\tau^{m}$. Then the prediction token $a_{t}$ (for example, $a_{t}$ can be ``\texttt{Barack}''\footnote{The names generated here may consist of more than one token; for instance, generating ``Barack'' results in the next token being ``Bar''. For ease of understanding, we will directly represent the next token as the corresponding complete word in this paper.}, ``\texttt{George}'' and other tokens) will be sampled for the distribution of $\textbf{P}(\bm{Y}|\bm{q},\bm{a}_{t-1},\mathcal{M})$:
\begin{equation}
    a_t \sim \textbf{P}(\bm{Y}|\bm{q},\bm{a}_{t-1},\mathcal{M}),
\end{equation}
where $a_t \in \bm{Y}$, and the sampling probability satisfies $p(a_t=\tau^{m})={p}(\tau^{m}|\bm{q},\bm{a}_{t-1},\mathcal{M})$ (e.g., the probability of next token being ``\texttt{Barack}'' is $0.377$ as shown in Fig.~\ref{fig:probability}).

\subsection{Failures in Traditional Uncertainty Estimation}

Traditional discriminative models typically use probability to estimate reliability. Probability, in fact, is a normalization of the strength of evidence for different categories. Due to the \emph{mutually exclusive} properties of the different categories, the relative strength between different categories can accurately indicate the reliability of the prediction. However, in the case of LLMs, the situation is different. Although LLM's next token prediction can still be viewed as a classification task with $|\bm{Y}|$ categories, these categories are no longer mutually exclusive. Even different tokens that are mutually exclusive in a conversation may no longer be mutually exclusive in a different context. This is why recent LLM research suggests shifting from token-to-token training to a concept-to-concept training paradigm. There may be more than one suitable next token, so the relative relationship between different tokens to no longer reflect the reliability of the response. Therefore, the information of the strength of the evidence before normalization becomes important. During the LLM training process, the suitable token accumulates evidence (logits increase, similar to the discriminative model~\cite{wei2022mitigating}), specifically, the higher the strength of the evidence, the more similar scenarios the model has encountered during training.

Consider the two situations shown in Fig.~\ref{fig:softmax}: (1) left: LLM has encountered this question 3 times during training, with answers a, b, and c; (2) right: LLM has encountered this question 3,000 times during training, and the answers can be summarized into three situations, a, b, and c. 

The accumulation of evidence in these two scenarios is significantly different, and their reliability is completely different. However, after logits are normalized into probabilities, the strength information of the evidence is lost, so in probability-based uncertainty estimation, these two situations are considered the same. This is because probability loses the key information that can indicate the reliability of the response.

\subsection{LogTokU: Four-quadrant Framework}\label{sec:four}

The reason why probability-based methods fail to identify reliability is that probability is normalized. After logits are normalized, only the relative strength between different answers (should be either ``\texttt{Barack}'' or ``\texttt{George}'') is retained, while the original strength information of the logits is discarded, which results in the loss of the ability to indicate reliability (distinguish between ``I do not know'' and ``I know more than one answer''). To address this limitation, we propose a Logits-induced Token Uncertainty framework termed LogTokU. In addition to considering the relative relationships among tokens (AU), LogTokU also takes into account the strength of the model's response (EU). With the information of EU, ``I do not know'' and ``I know more than one answer'' can be characterized separately. As shown in Fig.~\ref{fig:cover1}, the four quadrants of uncertainty are described as:

\textbf{Quadrant I: high AU, high EU}.\quad
This quadrant indicates that the model exhibits a low strength of evidence for all tokens, potentially due to lack of relevant knowledge. For example, as shown in Fig.~\ref{fig:cover2}, the model might recommend an unfamiliar medication.

\textbf{Quadrant II: low AU, high EU}.\quad
In this quadrant, the model shows a low evidence strength for most tokens but there is a higher strength for one particular token, indicating a lack of diversity despite producing a relatively high probability token. For example, the model repetitively suggests a drug name that was recently mentioned.
\emph{\textbf{Failure of probability-based methods:} Probability-based methods may regard this quadrant as highly reliable. However, it still involves a certain degree of risk due to the lack of knowledge, and the high probability only indicates the model recommendation.}

\textbf{Quadrant III: low AU, low EU}.\quad
Here, the model exhibits very high strength for one specific token while maintaining a low strength of evidence for all other tokens. This reflects a strong certainty about a particular token, such as the fixed phrase ``has been''.

\textbf{Quadrant IV: high AU, low EU}.\quad
This quadrant indicates that the model assigns a high strength of evidence to multiple tokens. Although none of them achieves a high probability, these tokens collectively demonstrate strong evidence. For example, this situation might arise when predicting nouns or punctuation marks that can be expressed in multiple valid ways.
\emph{\textbf{Failure of probability-based methods:} Probability-based methods may interpret this quadrant as unreliable. However, the model intends to express that there is more than one suitable candidate for the next token.}

\subsection{Considering Logits as Evidence}
In this subsection, we present a viable implementation of the LogTokU framework, specifically by modeling AU and EU using a Dirichlet distribution. Inspired by Evidential Deep Learning~\cite{sensoy2018evidential}, we treat logits as evidence for each token and model them into a Dirichlet distribution. A naive approach is using the logits of all non-negative values as evidence, while setting those with negative values to no evidence by applying the ReLU activation function. However, unlike conventional classification networks, the number of candidates generated by an LLM (that is, tokenizer size) is significantly large, with a considerable proportion of tokens having extremely low logits, and these tokens should be discarded as noise~\cite{tang2024top}. Therefore, this paper focuses on the distribution of main candidates with high logits. Specifically, we select the logits of tokens with the top $K$ largest logits to model a Dirichlet distribution:
\begin{equation}
    \alpha_k=\mathcal{M}({{\tau_k}}|\bm{q},\bm{a}_{t-1}), \quad \alpha_{0}=\sum_{k=1}^{K}\alpha_k,
    \label{eq:evidence}
\end{equation}
where \(\tau_{k}\) is the token with the $k$-largest prediction logit, and $\alpha_0$ is the total evidence of the Dirichlet distribution (the sum of the largest $k$ logits).

\textbf{Aleatoric (data) uncertainty.}\quad To measure the data uncertainty, we evaluate the expected entropy of the data distribution. Since entropy captures the ``peakiness'' of the output distribution, a lower entropy indicates that the model concentrates most of the probability mass on a single class, while a higher entropy characterizes a more uniform distribution, indicating that the model is undecided about the prediction. For Dirichlet networks, this quantity has a closed-form solution:
\begin{equation}
    \text{AU}(a_t) = - \sum_{k=1}^K\frac{\alpha_k}{\alpha_0}\left(\psi(\alpha_k+1) -  \psi(\alpha_0+1)\right), 
\end{equation}
where $\psi$ denotes the digamma function, defined as $\psi(x) = \frac{d}{d x} \log \Gamma(x)$. 

\textbf{Epistemic (model) uncertainty.}\quad 
We define the EU by:
\begin{equation}
    \text{EU}(a_t) = K / \sum_{k=1}^K (\alpha_k + 1),\label{eq:eu}
\end{equation}
and the underlying intuition is that larger $\alpha_k$ produces a sharper density, and thus it indicates increased confidence in a prediction. For more information on Dirichlet distribution, please refer to the introduction from~\cite{ulmer2023prior}.

\subsection{Theoretical Analysis: Why Logits can Capture Uncertainty}

Intuitively, the cross-entropy loss is not affected by the scale of logits; adding a constant to all elements of the logits vector does not change the probability values. For instance, both [20, 18] and [30, 28] are mapped to the probability distribution [0.88, 0.12] after applying the softmax function. However, intuition is not always trustworthy. During the actual gradient descent training of models, the scale of logits is constrained. The cross-entropy loss can be decomposed into a format similar to the evidential deep learning loss function, which naturally penalizes shifting logits scales during gradient updates:
\begin{equation}
\begin{aligned}
    \mathcal{L}_\text{LLM} &=  \underbrace{-\left( \sum_{j}^{|\bm{Y}|} y_{j}\log \frac{z_{j}+1}{\sum_{j'}^{|\bm{Y}|} (z_{j'}+1)} \right)}_{\text{Evidential deep learning classification term}} \\& \quad - \sum_{j}^{|\bm{Y}|} y_{j}((z_{j}+1) - \log (z_{j}+1)) \\& \underbrace{- \log \frac{\sum_{j'}^{|\bm{Y}|} (z_{j'}+1)}{\sum_{j'}^{|\bm{Y}|} e^{(z_{j'}+1)}} }_{\text{Evidence regularization}},\label{eq:loss_cp}
\end{aligned}
\end{equation}
which is very close to the format of training loss in evidential modeling shown in the Appendix~\ref{app:evd}, and the following analysis show the detailed differences and corresponding impacts between evidential modeling loss and CE loss.

\textit{Bridging the positivity gap.} A critical distinction lies in evidence parameterization: evidential models enforce $z_j \geq 0$ through architectural constraints (e.g., ReLU activations), whereas standard CE allows unbounded logits. However, the exponential transformation in softmax ($e^{z_j}$) renders negative logits equivalent to weak evidence (values near 0), while positive logits represent strong evidence. Thus, both frameworks ultimately operate in a non-negative evidence space, with CE's logits implicitly encoding $\log(\text{evidence} + \epsilon)$ where $\epsilon \to 0$. This logarithmic relationship preserves the uncertainty-quantification mechanism while relaxing explicit positivity constraints.

\textit{The role of gradient dynamics.} The purported scale invariance of CE holds only for completed training – during optimization, gradient updates actively regulate logit magnitudes. For misclassified samples, the $\log\sum e^{z_j}$ term exerts downward pressure on all logits, while the $-y_j z_j$ term selectively boosts the true class logit. This creates an implicit evidence threshold: logits only grow when the model can simultaneously increase the true class evidence while suppressing competing terms, mirroring the explicit trade-off in evidential learning.

\section{Application I: Dynamic Decoding Strategy}\label{sec:case1}

\subsection{LogTokU-guided Decoding}
It is necessary to ensure that the generated response to be of diversity, especially in diversity-driven fields such as LLM-guided discovery researches~\cite{peng2024large}. However, higher diversity often means that the LLM's responses are more prone to hallucinations. For example, during the sampling process, a larger temperature tends to generate more unreliable responses from the model. Therefore, ensuring both the diversity and precision of LLM-generated results is a key challenge.

LogTokU offers a potential solution to this challenge. In this subsection, we propose a dynamic decoding strategy that can adjust its sampling approach according to LogTokU during response generation. The dynamic decoding strategy ensures diverse answers when the LLM has adequate knowledge (i.e., low model uncertainty), while adopting a more cautious sampling strategy when the model's knowledge is insufficient (i.e., high model uncertainty), thus maintaining both diversity and accuracy in the generated responses. Specifically, we hope that the sampling diversity and the LLM's EU about the next token denoted as $\text{EU}(a_t)$ are negatively correlated,
which means that the higher EU, the less chance for sampling tokens with lower scores.
Taking the temperature sampling strategy as an example, there is a smaller temperature when the model has larger EU.

\subsection{Experimental Analysis}
\subsubsection{Settings}
\begin{figure}
\centering 
\includegraphics[width=0.8\textwidth]{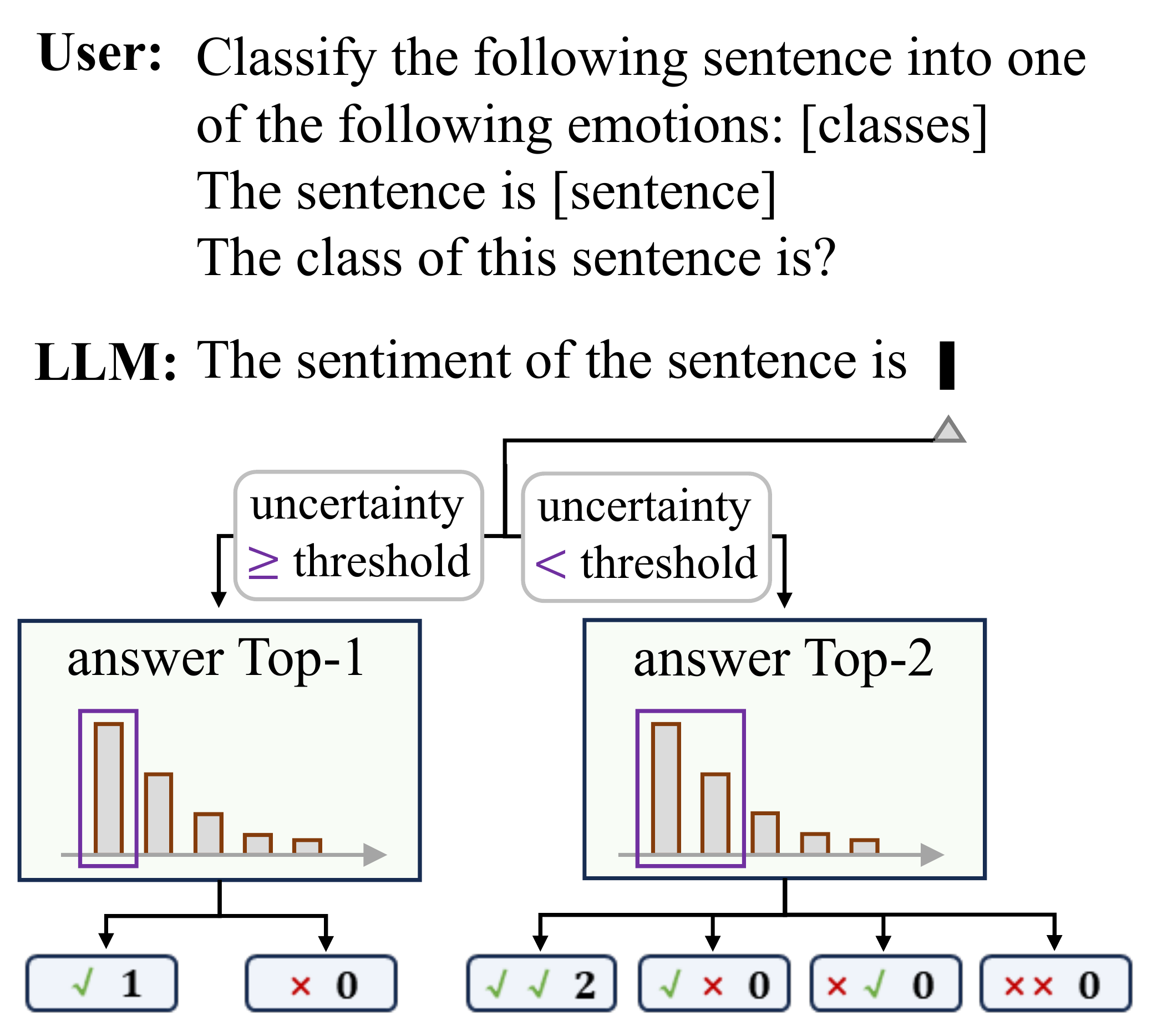} 
\caption{Illustration of experimental setting in Table~\ref{tab:percent}.} \label{fig:seting}
\end{figure}

In this paper, we use the multi-label evaluation benchmark SemEval~\cite{mohammad-etal-2018-semeval}, which is a multi-tag NLP analysis task on tweet text. We evaluate models of different sizes, including LLaMA2 (7B)\footnote{\href{https://huggingface.co/meta-llama/Llama-2-7b-chat-hf}{https://huggingface.co/meta-llama/}}, LLaMA2 (13B), LLaMA3 (3B), LLaMA3 (8B) and LLaMA3 (70B). As shown in Fig.~\ref{fig:seting}, after providing an answer with the first class label, LLM dynamically decides whether to give the second class label based on the uncertainty indicator. We can evaluate the ability of uncertainty to guide the model in rejecting incorrect answers (the ability to avoid hallucinations in generating diverse responses).

\begin{table*}
\centering
\caption{\textbf{Dynamic Decoding Performance Comparison.}\quad The $\underline{scores}$ indicate the best performance of the compared methods and $\textcolor{mycolor2}{{\Uparrow}_{rate\%}}$ represents the improvement relative to the best of the compared methods.}
\label{tab:percent}
\resizebox{\textwidth}{!}{ 
\begin{tabular}{c|ccc|ccc}
\toprule

{\textbf{Method}}&  LLaMA2$_{\text{(7B)}}$ & LLaMA2$_{\text{(13B)}}$& LLaMA2$_{\text{(70B)}}$ & LLaMA3$_{\text{(3B)}}$& LLaMA3$_{\text{(8B)}}$ & LLaMA3$_{\text{(70B)}}$    \\
\midrule
\multicolumn{1}{c|}{Greedy Search} &$77.48\%$&$74.19\%$&\underline{$69.96\%$}&$71.06\%$&$79.10\%$&$77.17\%$\\
\multicolumn{1}{c|}{Top-2} &$77.32\%$&$73.70\%$&$64.31\%$&$79.35\%$&$91.87\%$&$83.09\%$\\
\multicolumn{1}{c|}{Probability} &$78.49\%$&$74.19\%$&$69.93\%$&$81.10\%$&$93.07\%$&$83.40\%$\\
\multicolumn{1}{c|}{Entropy} &\underline{$79.32\%$}&\underline{$74.87\%$}&$69.93\%$&\underline{$82.33\%$}&\underline{$94.75\%$}&\underline{$84.01\%$}\\
 \midrule
\multicolumn{1}{c|}{LogTokU} &\improved{85.87}{6.6}&\improved{86.22}{11.4}&\improved{76.19}{6.2}&\improved{83.92}{1.6}&\improved{97.55}{2.8}&\improved{89.11}{5.1}\\
\bottomrule
\end{tabular}} 
\end{table*}

To evaluate the effectiveness of the dynamic decoding strategy, we verify whether it can enhance diversity while maintaining accuracy. Specifically, we evaluate whether the model can select as many correct answers as possible with the guidance of uncertainty on the multi-label LLM benchmark SemEval~\cite{mohammad-etal-2018-semeval}. As shown in Fig.~\ref{fig:seting}, for all test samples in the entire data set, when generating responses, the LLM selects the class with the highest output probability or the two tokens with the highest probabilities at the critical token (class) position based on uncertainty. The rule is similar to multiple-choice questions in exams. For any question, the LLM can decide whether to answer a second class. If it answers, there are two possibilities: (1) the second class is correct, the LLM will get one more point for its diversity; (2) the second class is incorrect, which may result in losing the points earned from the first correct answer. In other words, the LLM needs to estimate its own confidence and balance the trade-off between the penalty (hallucination) and award (diversity).


\textbf{Compared methods}. We introduce two baseline decoding strategies and two dynamic decoding strategies based on probability and entropy, including: $\bullet$ Greedy Search: the LLM sacrifices diversity and selects only one class for all samples with the highest probability. $\bullet$ Top-2: the LLM seeks diversity and selects two classes for all samples with the highest probability and the second highest probability. $\bullet$ Probability: the LLM dynamically chooses to select either one class or two classes for different samples, with the maximum probability as the indicator. The LLM selects one class when the maximum probability is low and two classes when the maximum probability is high. $\bullet$ Entropy: This is also a dynamic strategy with entropy as an indicator. The LLM selects one class when the entropy is low and two classes when the entropy is high. $\bullet$ LogTokU: This is also a dynamic strategy, with the indicator being $\text{EU}(a_t)$, as shown in Sec.~\ref{sec:case1}. The LLM selects one class when EU is high and two classes when EU is low.

\begin{figure}
\centering
\includegraphics[width=0.8\textwidth]{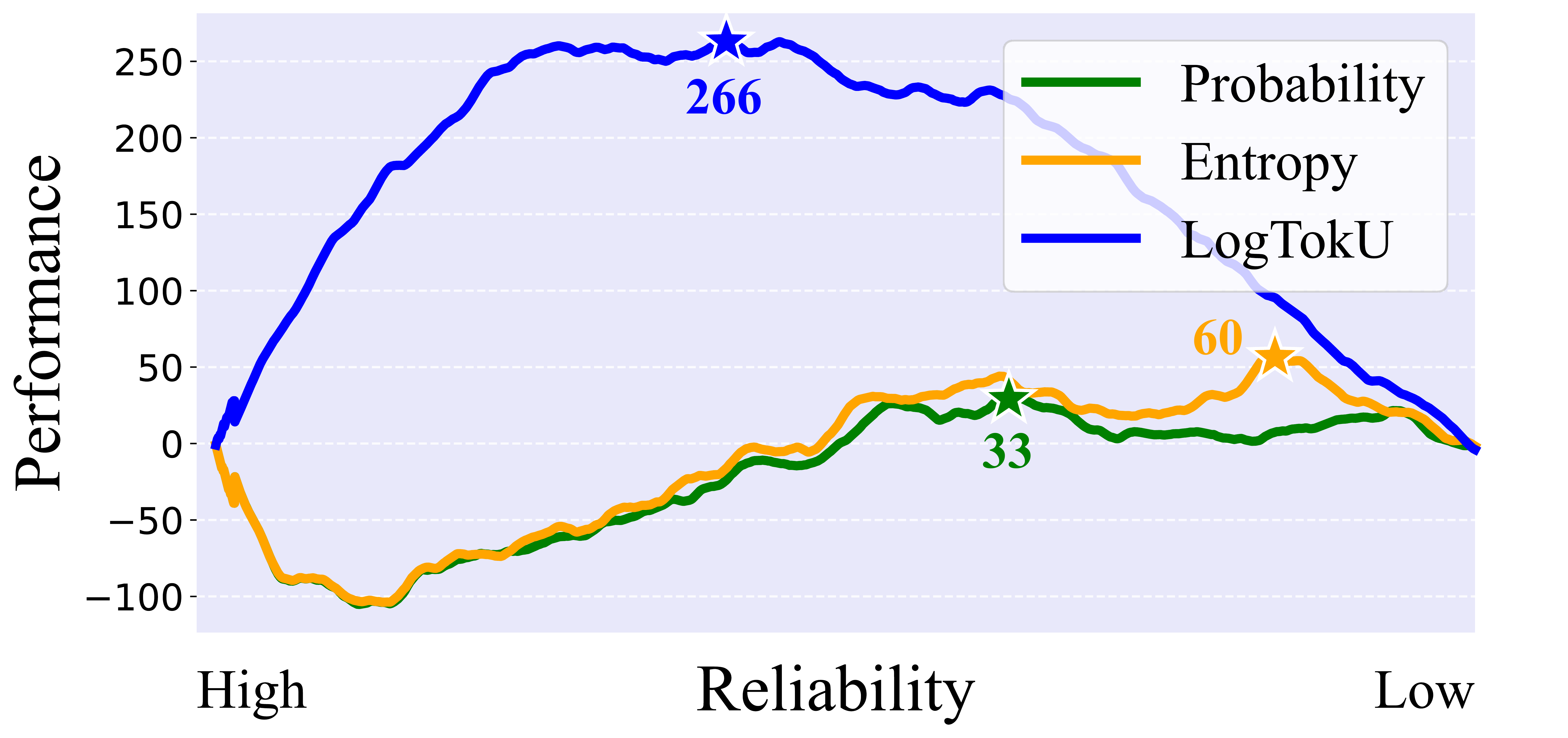}
\caption{A close-up observation explains why LogTokU achieves the best performance. All samples are sorted by reliability from high to low. The ``performance'' is the accumulated score (i.e., the number of accumulated correct responses minus the number of accumulated incorrect responses). A trend of increasing and then decreasing represents a good reliability indicator, where the answer becomes more likely to be wrong as reliability decreases.}
\label{fig:curve}
\end{figure}

\subsubsection{Results Analyses}

We show the normalized performance of different methods in various LLM sizes (higher is better), as shown in Table~\ref{tab:percent}. We can find that using LogTokU achieves the best performance, with a significant improvement in the overall score of the model, with an improvement of more than 10\% on LLaMA2-13B. 

\begin{table*}
\centering
\caption{\textbf{Response Reliability Estimation Performance Comparison.}\quad The ${auroc}\%$ scores indicate the performance of the relationship between the estimated reliability and response correctness (BLEURT$>0.5$) on TruthfulQA benchmark of the compared methods, $O(1)$ indicates the complexity in terms of response sampling (be free of multiple sampling or not), and $\textcolor{mycolor2}{{\Uparrow}_{rate\%}}$ indicates the improvement relative to the best performance of all compared methods.}
\label{tab:reliability}
\resizebox{\textwidth}{!}{ 
\begin{tabular}{c|c|ccc|ccc}
\toprule
{\textbf{Method}}& $O(1)$ &  LLaMA2$_{\text{(7B)}}$ & LLaMA2$_{\text{(13B)}}$& LLaMA2$_{\text{(70B)}}$ & LLaMA3$_{\text{(3B)}}$& LLaMA3$_{\text{(8B)}}$ & LLaMA3$_{\text{(70B)}}$    \\
\midrule


\multicolumn{1}{c|}{Probability} &\checkmark&$61.39\%$&$68.38\%$ &$66.38\%$&$66.71\%$&$64.78\%$&$61.15\%$\\
\multicolumn{1}{c|}{Entropy}& \checkmark &{$64.30\%$}&{$72.53\%$}&{$69.06\%$}&{$68.64\%$}&{$66.50\%$}&{$63.96\%$}\\
\cmidrule(lr){2-2} \cmidrule(lr){3-5} \cmidrule(lr){6-8}
\multicolumn{1}{c|}{LN-E~\cite{malinin2021uncertainty}} &\xmark &$41.15\%$&$44.07\%$&$43.03\%$&$44.60\%$&$44.95\%$&$47.74\%$\\
\multicolumn{1}{c|}{SE~\cite{kuhn2023semantic}} & \xmark&$49.20\%$&$53.71\%$&$49.47\%$&$57.25\%$&$53.98\%$&$62.91\%$\\
\multicolumn{1}{c|}{DSE~\cite{kuhn2023semantic}} &\xmark&$49.36\%$&$53.97\%$&$49.76\%$&$57.78\%$&$54.56\%$&$62.85\%$\\
\multicolumn{1}{c|}{LeS~\cite{lin2024generating}} &\xmark &$55.92\%$ &$56.83\%$ &$57.09\%$ &$60.90\%$ &$58.85\%$ &$60.76\%$\\

\multicolumn{1}{c|}{Average} & -&$48.91\%$&$52.15\%$&$49.84\%$&$55.13\%$&$53.09\%$&$58.57\%$\\ 
 \midrule
\multicolumn{1}{c|}{LogTokU}&\checkmark &
     \relimproved{71}{78}{7.5}
&\relimproved{79}{78}{7.3}&\relimproved{71}{83}{2.8}&\relimproved{71}{03}{2.4}&\relimproved{73}{63}{7.1}&\relimproved{69}{67}{5.7}\\
\bottomrule
\end{tabular}} 
\end{table*}

As shown in Fig.~\ref{fig:seting}, when the model chooses to generate more than one answer, the score obtained may either increase (award of correct response) or decrease (penalty of error response), and a trustworthy response reliability indicator should have the ability to identify the error responses. Therefore, we estimate the accumulated score curve as LLM answers more and more questions with diverse answers. Specifically, we arrange all samples in ascending order of estimated uncertainty of all methods, then the samples with uncertainty lower than threshold will answer more than one answer, which the others just answer the choice with the highest prediction probability.
Ideally, the accumulated score change should follow a trend that first increases and then decreases, which implies that samples with lower uncertainty are more likely to gain additional score when challenged to give more diverse responses. As the uncertainty increases, the probability of giving the response to be error also increases. The results are shown in Fig.~\ref{fig:curve}, and it can be observed that the uncertainty indicated by LogTokU consistently satisfies the expectation. However, traditional probability-based uncertainty estimation methods show a decrease trend when the estimated reliability is high, indicating that these methods fail to estimate the risk of giving more diverse responses. We highlight the best performance points in terms of accumulated score for different methods, and it can be observed that our method significantly outperforms the comparison methods.

\section{Application II: Reliability Estimation}
\subsection{LogTokU-guided Response Uncertainty Estimation}

Another benefit that LogTokU brings is the estimation of the reliability of the response. In traditional probability-based token uncertainty estimation, a large number of uncritical tokens exhibit high uncertainty, making it difficult to map token uncertainty to sentence uncertainty. This issue requires manually applying weights to different tokens to overcome the problem caused by uncritical tokens being estimated with high uncertainty. LogTokU naturally overcomes this problem. As shown in the case study in Fig.~\ref{fig:cover2}, for uncritical tokens such as commas, the probability-based method labels them as unreliable tokens due to their low predictive probability (high entropy). In contrast, LogTokU accurately classifies them into the fourth quadrant, marking them as ``I know more than one answer''. Therefore, LogTokU can more easily use token uncertainty to represent the uncertainty of sentences. Inspired by~\cite{duan2024shifting}, we use the most uncertain tokens in a sentence to represent the overall reliability of the sentence. The response reliability can be represented as the averaged reliability on tokens with the $K$-lowest reliability values:
\begin{equation}
    \mathcal{R}_{\text{response}} = \frac{1}{K} \sum_{t \in \mathcal{T}_{K}} \mathcal{R}(a_{t}),
\end{equation}
where $ \mathcal{R}_{\text{response}}$ indicates the reliability of the response and $\mathcal{R}(a_{t})$ represents the reliability of token $a_t$, and \(\mathcal{T}_{K}\) represents the set of \(K\) tokens with the lowest reliability values (\(\text{rel}(a_{t})\)).

Following the uncertainty combination of AU and EU in discriminative models~\cite{abdar2021review}, in the experiments, we simply represent the reliability of the token as: 
\begin{equation}
    \mathcal{R}(a_{t}) =-{\text{AU}(a_t)\cdot \text{EU}(a_t)},
\end{equation}
which indicates that the reliability of the token is low when both the estimated AU and EU are high. In this paper, we simply identify the response reliability in QA task according to the scenarios classified in Fig.~\ref{fig:cover1}: (1) high AU, high EU: LLM lacks knowledge of the question and has no idea of a suggested answer (unreliable); (2) low AU, high EU: The LLM lacks knowledge of the question, but it knows what should be an appropriate answer (reliable); (3) low AU, low EU: LLM knows precisely what is the most appropriate answer (reliable);
(4) high AU, low EU: LLM has encountered many similar samples during training and knows more than one suitable response (reliable).

\subsection{Experimental Analysis}
\subsubsection{Settings}
To validate whether the estimated response reliability is trustworthy, we evaluate it on the TruthfulQA benchmark dataset~\cite{lin2021truthfulqa}. Following standard settings, we consider responses with $\text{BLEURT} > 0.5$ as correct answers, and use the estimated reliability as the score for the responses to calculate the AUROC. A higher AUROC indicates that responses estimated with lower reliability are more likely to be incorrect. We compare LogTokU with probability-based strategy, entropy-based strategy and recent sampling-based uncertainty estimation methods, including: LN-Entropy (LN-E)~\cite{malinin2021uncertainty}, Semantic Entropy (SE)~\cite{kuhn2023semantic}, D-Semantic Entropy (DSE)~\cite{kuhn2023semantic}, Lexical Similarity (LeS)~\cite{lin2024generating}. Specifically, the token reliability based on probability is represented as: $\mathcal{R}(a_t)=\log \left(p(a_t|\bm{q},\bm{a}_{t-1},\mathcal{M})\right)$, and the token reliability based on entropy is represented as: $\mathcal{R}(a_t)=\frac{1}{H(a_t)}$, where $H(a_t)$ indicates the entropy of prediction distribution of $a_t$.

\subsubsection{Results Analyses}

Table~\ref{tab:reliability} shows the AUROC performance of different estimation methods in the TruthfulQA benchmark, with a higher value indicating a more reliable response reliability estimation. The results show that LogTokU achieves the best performance in models of various sizes, suggesting that LogTokU provides a better estimation of the response reliability. Furthermore, we find that sampling methods do not perform well on this task, and similar observations have been reported in~\cite{xiong2024efficient}. To illustrate the performance gap between sampling-based methods and LogTokU, we calculate the average performance of current sampling-based methods termed ``Average''. One potential reason is because sampling-based methods cannot capture the situation in the Quadrant IV and thus fail to measure models' inherent uncertainty. For more experimental results, please refer to the appendix.

\section{Conclusion}
In this paper, we find that the main reason for the failure of probability-based methods in estimating LLM's uncertainty is that the information regarding evidence strength is lost during the normalization process, making probability cannot accurately reflect reliability. Therefore, we propose LogTokU to model the evidence for generating the next token in LLMs, a real-time token uncertainty estimation framework. We use LogTokU to guide two downstream tasks of LLMs, dynamic decoding and response reliability estimation, achieving significant performance improvements. These downstream tasks demonstrate the simplicity, effectiveness, and immense potential of LogTokU.


%

\ifCLASSOPTIONcompsoc
  \section*{Acknowledgments}
\else
  \section*{Acknowledgment}
\fi

{
\bibliographystyle{IEEEtran}
\bibliography{bare_jrnl_compsoc}
}

\ifCLASSOPTIONcaptionsoff
  \newpage
\fi

\clearpage
\onecolumn
\appendices

\clearpage
\section{Word Uncertainty}
The uncertainty estimation of LogTokU indicates which of the four conditions applies to each token generated by LLM. However, many LLMs encode words based on their root forms, for example, ``\texttt{positive}'' is embedded as ``\texttt{pos}'' and ``\texttt{itive}'' in LLaMA\footnote{Token id: pos\_1066, itive\_3321, please note that the word ``positive'' referred here is different from another word ``\_positive''.}, making token-level uncertainty does not necessarily correspond to word-level uncertainty. In this section, we propose an approach to visualize the word-level uncertainty estimated by LogTokU. Specifically, we represent the uncertainty of a word as the maximum uncertainty among the tokens that constitute the word:
\begin{equation}
\begin{aligned}
    &\text{AU}(\texttt{word})=\max_{\alpha_t \in \texttt{word}}(\text{AU}(\bm{\alpha}_t)),\\
    &\text{EU}(\texttt{word})=\max_{\alpha_t \in \texttt{word}}(\text{EU}(\bm{\alpha}_t)).
\end{aligned}
\end{equation}

To emphasize the contrast, in Fig.~\ref{fig:cover2}, we set the AU and EU of tokens with uncertainty lower than the mean of the entire response to zero and normalize the values to the range $[0,1]$. At the same time, we represent unreliability as:
\begin{equation}
    \text{unrel}(\texttt{word})={\text{AU}(\texttt{word})\times\text{EU}(\texttt{word})},
\end{equation}
which implies that when both AU and EU are high (quadrant IV), the reliability of the word is the lowest.

Such a tool can significantly improve the explainability of LLM responses, particularly in QA quadrants such as medical applications, thus enhancing the experience and efficiency of the human-LLM interaction.

\section{Detailed Theoretical Analysis}
\subsection{Different correct answers are competitor}

In this subsection, we will show the details about the derivation process. For any LLM trained with cross-entropy loss, different correct answers are competitors in terms of probability. Continuing with the example of proposing a president, suppose $\tau^{a}$ (``\texttt{Barack}'') is the label of a sample whose $\bm{q}$ is ``\texttt{[INST]Could you give me one name of president?[\textbackslash INST]}'' and a generated token vector $\bm{a}_{t-1}$  can be decoded into ``\texttt{Sure, here is a historical American president:**}'', the loss of the next token at this position during supervised fine-tuning can be written as:
\begin{equation}
\begin{aligned}
 &L^{\tau^a} = - \log \frac{\exp(\mathcal{M}({\tau^a}|\bm{q},\bm{a}_{t-1}))}{\sum_{m=1}^{|\bm{Y}|} \exp(\mathcal{M}(\tau^{m}|\bm{q},\bm{a}_{t-1}))} ,
 \\   &L^{\tau^b} = - \log \frac{\exp(\mathcal{M}(\tau^b|\bm{q},\bm{a}_{t-1}))}{\sum_{m=1}^{|\bm{Y}|} \exp(\mathcal{M}(\tau^{m}|\bm{q},\bm{a}_{t-1}))} ,
\end{aligned}
\end{equation}
where $L^{\tau^a}$ is the loss on the sample with the next token label $\tau^{a}$.
Consider cases where multiple distinct answers to the same question appear in the training set, the situation becomes different. For example, $\tau^{b}$ (``\texttt{George}'') is the label in another sample with the same question \footnote{The ``same question'' refers to questions that are semantically equivalent but do not need to be identical.}. When the model is simultaneously fine-tuned on both samples, the gradient update for the model will be:
\begin{equation}
\begin{aligned}
 & \nabla_{\mathcal{M}} (L^{\tau^a} + L^{\tau^b}) = \nabla_{\mathcal{M}} L^{\tau^a} + \nabla_{\mathcal{M}} L^{\tau^b} \\
&= -y_a^{\tau^a}\frac{1}{\Omega_a^{\tau^a}}\nabla_{\mathcal{M}}\Omega_a^{\tau^a}-\sum_{m \neq a}^{|\bm{Y}|} y_a^{\tau^m}\frac{1}{\Omega_a^{\tau^m}}\nabla_{\mathcal{M}}\Omega_a^{\tau^m}
\\
&\quad -y_b^{\tau^b}\frac{1}{\Omega_b^{\tau^b}}\nabla_{\mathcal{M}}\Omega_b^{\tau^b}-\sum_{m \neq b}^{|\bm{Y}|} y_b^{\tau^m}\frac{1}{\Omega_b^{\tau^m}}\nabla_{\mathcal{M}}\Omega_b^{\tau^m}
\\
&= \underbrace{-y_a^{\tau^a}\frac{1}{\Omega_a^{\tau^a}}\nabla_{\mathcal{M}}\Omega_a^{\tau^a}-y_b^{\tau^b}\frac{1}{\Omega_b^{\tau^b}}\nabla_{\mathcal{M}}\Omega_b^{\tau^b}}_{\text{(1) maximizing the probability of annotated answer}}\\& \quad \underbrace{-y_a^{\tau^b}\frac{1}{\Omega_a^{\tau^b}}\nabla_{\mathcal{M}}\Omega_a^{\tau^b}-y_b^{\tau^a}\frac{1}{\Omega_b^{\tau^a}}\nabla_{\mathcal{M}}\Omega_b^{\tau^a}}_{{\text{\textbf{(2)} minimizing the probability of the other annotated answer}}}\\& \quad \underbrace{-\sum_{m \neq a,b}^{|\bm{Y}|}y_{a,b}^{\tau^m} \left[ \frac{1}{\Omega_a^{\tau^m}}\nabla_{\mathcal{M}}\Omega_a^{\tau^m} + \frac{1}{\Omega_b^{\tau^m}}\nabla_{\mathcal{M}}\Omega_b^{\tau^m} \right]}_{\text{(3) minimizing the probability of incorrect answers}},\label{eq:competitor}
\end{aligned}
\end{equation}
where $\Omega_a^{\tau^a}=\frac{\exp(\mathcal{M}(\tau^a|\bm{q},\bm{a}_{t-1}))}{\sum_{m=1}^{|\bm{Y}|} \exp(\mathcal{M}(\tau^{m}|\bm{q},\bm{a}_{t-1}))}$, and $y_a^{\tau^m}$ indicates the next token label of a training sample with ground-truth label ${\tau^a}$, that is, we have $y_a^{\tau^a}=1$ and $y_a^{\tau^b}=0$. In particular, when $\mathcal{M}$ is in a certain state during training, we have $\Omega_a^{\tau^a}=\Omega_b^{\tau^a}$, and we make distinctions to facilitate the reader's understanding here. As we can see, for scenarios with multiple answers, the training objective can be divided into three parts:

(1) For each sample, increase the probability of its own annotation in the output distribution. 

$\bullet$ For example, a sample labeled ``\texttt{Barack}'' encourages the model to predict the next token as ``\texttt{Barack}'' with higher probability;

(2) For each sample, decrease the probability of another sample's annotation in the output distribution.

$\bullet$ For example, a sample labeled ``\texttt{Barack}'' encourages the model to minimize the probability of predicting the next token as ``\texttt{George}''; \textit{\textbf{Note:}} This part leads to the issue where probability can no longer capture the reliability of LLM responses, as different correct answers tend to reduce the probability of other correct answers, making low probabilities unreliable indicators of high uncertainty.

(3) For both samples, decrease the probability of other outputs not present in the annotations in the output distribution. 

$\bullet$ For example, a sample labeled ``\texttt{Barack}'' and a sample labeled ``\texttt{George}'' both encourage the model to minimize the probability of predicting the next token as ``\texttt{Coffee}''.

\subsection{Max Token Probability Can \textit{Not} Represent Risk of A Wrong Answer}
As shown in Eq.~\ref{eq:competitor}, the part (2) harms the confidence (probability) of correct predictions when there is more than one correct answer during training. This phenomenon can also be analyzed from the perspective of Mixup~\cite{zhang2017mixup}, the target for a sample with more than one correct answer will be a mix target across all labeled answers. This means that the more answers in the training data for a question, the lower the maximum probability of every single correct answer. During training, suppose $\tau^{a}$ and $\tau^{b}$ are both annotation labeled answers for $\bm{q}$, the expected confidence of a probability-based indicator is
\begin{equation}
p(\tau^{a}|\bm{q},\bm{a}_{t-1},\mathcal{M})=p(\tau^{b}|\bm{q},\bm{a}_{t-1},\mathcal{M}) = 1,
\end{equation}
however, what we actually have is 
\begin{equation}
\sum_{m=1}^{|\bm{Y}|}p(\tau^{m}|\bm{q},\bm{a}_{t-1},\mathcal{M}) = 1,\label{eq:mixup}
\end{equation}
which means that $\tau^{a}$ and $\tau^{b}$ share a total probability less than $1$, leading to that correct tokens may not have large probabilities. Consequently, even when we assign varying levels of importance to different words and assign a higher weight to ``critical tokens'', the probability may not reflect the reliability of the response, such as the case shown in Fig.~\ref{fig:probability}.

\subsection{How can Evidential Modeling Capture Uncertainty} \label{app:evd}

Due to the absence of ground-truth uncertainty, uncertainty-aware training methods like evidential deep learning~\cite{sensoy2018evidential} do not directly supervise EU. Instead, they implicitly learn to quantify uncertainty by imposing inductive biases during optimization. This is achieved through a dual mechanism: (1) a classification term that aligns predictions with labels, and (2) a regularization term that constrains evidence accumulation. Specifically, the model is penalized for accumulating excessive evidence (large logits) unless it achieves high confidence in predictions. The training loss $\mathcal{L}_\text{EVD}$ exemplifies this principle:

\begin{equation}
\begin{aligned}
    \mathcal{L}_\text{EVD} =\underbrace{-\left( \sum_{j}^{|\bm{Y}|} y_{j}\log \frac{z_{j}+1}{\sum_{j'}^{|\bm{Y}|} (z_{j'}+1)} \right)}_{\text{Classification term}} \underbrace{+ \mathcal{L}_{\text{reg}}}_{\text{Evidence regularization}},
\end{aligned}
\end{equation}

where $z_j$ represents non-negative evidence parameters for class $j$, and $\mathcal{L}_{\text{reg}}$ penalizes premature evidence accumulation.

\section{Detailed Experimental Results}
\subsection{Ablation on K}

\begin{table*}[th]
\centering
\caption{\textbf{Response reliability estimation performance under different $K$.} \quad Results are based on the \texttt{LLaMA2-chat-13b-hf}.The $K=\text{all}$ indicates that we use all the tokens to estimate the reliability, and $\textcolor{mycolor2}{{\Uparrow}_{rate\%}}$ indicates the improvement relative to the best performance of all compared methods.}
\resizebox{\textwidth}{!}{ 
\begin{tabular}{c|cccccc|c}
\toprule
{\textbf{Method}}&  $K=1$ & $K=5$&  $K=10$ &  $K=15$&  $K=20$& $K=25$& $ K=\text{all}$  \\
\midrule
\multicolumn{1}{c|}{Probability}&{$49.90\%$}&{$50.66\%$}&{$57.60\%$}&{$62.84\%$}&{$66.27\%$}&{$68.38\%$}&{$66.14\%$}\\
\multicolumn{1}{c|}{Entropy}&{$56.62\%$}&{$55.90\%$}&{$62.31\%$}&{$67.51\%$}&{$70.64\%$}&{$72.53\%$}&{$68.76\%$}\\
\multicolumn{1}{c|}{LogTokU}&
     \relimproved{65}{70}{9.1}
&\relimproved{66}{95}{11.1}&\relimproved{73}{84}{11.5}&\relimproved{77}{98}{10.5}&\relimproved{79}{59}{9.0}&\relimproved{79}{78}{7.3}&\relimproved{69}{08}{0.3}\\
\bottomrule
\end{tabular}} 
\end{table*}
\subsection{Result with LLM-as-Judge}

\begin{table*}[th]
\centering
\caption{\textbf{Response Reliability Estimation Performance Comparison.}\quad The ${auroc}\%$ scores indicate the performance of the relationship between the estimated reliability and response correctness (LLM-Judge$=1$) on TruthfulQA benchmark of the compared methods, $O(1)$ indicates the complexity in terms of response sampling (be free of multiple sampling or not), and $\textcolor{mycolor2}{{\Uparrow}_{rate\%}}$ indicates the improvement relative to the best performance of all compared methods.}
\resizebox{\textwidth}{!}{ 
\begin{tabular}{c|c|ccc|ccc}
\toprule
{\textbf{Method}}& $O(1)$ &  LLaMA2$_{\text{(7B)}}$ & LLaMA2$_{\text{(13B)}}$& LLaMA2$_{\text{(70B)}}$ & LLaMA3$_{\text{(3B)}}$& LLaMA3$_{\text{(8B)}}$ & LLaMA3$_{\text{(70B)}}$    \\
\midrule


\multicolumn{1}{c|}{Probability} &\checkmark&$53.18\%$&$55.89\%$ &$55.03\%$&$62.70\%$&$56.37\%$&$53.64\%$\\
\multicolumn{1}{c|}{Entropy}& \checkmark &{$55.34\%$}&{$58.95\%$}&{$56.26\%$}&{$64.76\%$}&{$58.33\%$}&{$55.26\%$}\\
\cmidrule(lr){2-2} \cmidrule(lr){3-5} \cmidrule(lr){6-8}
\multicolumn{1}{c|}{LN-E~\cite{malinin2021uncertainty}} &\xmark &$45.15\%$&$42.93\%$&$46.24\%$&$44.48\%$&$44.46\%$&$45.30\%$\\
\multicolumn{1}{c|}{SE~\cite{kuhn2023semantic}} & \xmark&$53.96\%$&$57.17\%$&$56.20\%$&$59.54\%$&$57.87\%$&$56.78\%$\\
\multicolumn{1}{c|}{DSE~\cite{kuhn2023semantic}} &\xmark&$54.18\%$&$57.10\%$&$56.72\%$&$59.89\%$&$57.94\%$&$56.65\%$\\
\multicolumn{1}{c|}{LeS~\cite{lin2024generating}} &\xmark &$52.57\%$ &$51.23\%$ &$53.99\%$ &$58.29\%$ &$55.95\%$ &$55.85\%$\\

\multicolumn{1}{c|}{Average} & -&$51.72\%$&$52.11\%$&$53.54\%$&$55.80\%$&$54.81\%$&$53.90\%$\\

 \midrule
\multicolumn{1}{c|}{LogTokU}&\checkmark &
     \relimproved{59}{07}{3.7} & \relimproved{61}{91}{3.0} & \relimproved{56}{70}{0.4} & \relimproved{66}{16}{1.4} & \relimproved{61}{10}{2.8} & \relimproved{61}{04}{5.8} \\
\bottomrule
\end{tabular}} 
\end{table*}
In our main paper, a generation is considered truthful when its BLUERT score exceeds a threshold of 0.5. In this ablation experiment, following \cite{xiong2024efficient}, we adopt a more contemporary and independent approach by utilizing \verb|Meta-Llama-3-8B-Instruct| to evaluate the correctness of the generated answers. We employ a 2-shot prompt for this purpose. The results presented here are aligned with those in Table 2 of the main paper.

\section{Implementation Details}
\subsection{Prompts for Different Experiments}
\begin{tcolorbox}[colback=white, colframe=black, coltitle=white, colbacktitle=black,
    title={Prompt for Response Reliability Estimation}, boxrule=0.5mm, sharp corners]

    \noindent \textbf{Following the previous work, we use the following prompts for the LLaMa2 and LLaMa3 series, respectively.}

    \vspace{10pt}

    \noindent \textbf{LLaMa2 Series Prompt} 
    
    \noindent \texttt{Answer the question concisely. Q: \{\textcolor{red}{question}\} A:}

    \vspace{10pt}

    \noindent \textbf{LLaMa3 Series Prompt}

    \noindent \texttt{<|start\_header\_id|>user<|end\_header\_id|>\textbackslash n\textbackslash n Answer the question concisely. Q: \{ \{\textcolor{red}{question}\} \} A:<|eot\_id|> assistant\textbackslash n\textbackslash n}

\end{tcolorbox}
\begin{tcolorbox}[colback=white, colframe=black, coltitle=white, colbacktitle=black,
    title={Prompt for Dynamic Decoding Strategy}, boxrule=0.5mm, sharp corners]

    \noindent \textbf{Following the recommendations from \href{https://huggingface.co/meta-llama/Llama-2-7b-chat-hf}{Hugging Face}, we use the following prompts for the LLaMa2 and LLaMa3 series, respectively.}

    \vspace{10pt}

    \noindent \textbf{LLaMa2 Series Prompt}

    \noindent \texttt{<s>[INST]Classify the following sentence into ['anger', 'anticipation', 'disgust', 'fear', 'joy', 'love', 'optimism', 'pessimism', 'sadness', 'surprise', 'trust'], the sentence is \{\textcolor{red}{sentence}\} [/INST] The class of this sentence is: \textbackslash n['}

    \vspace{10pt}

    \noindent \textbf{LLaMa3 Series Prompt}

    \noindent \texttt{<|begin\_of\_text|><|start\_header\_id|>system<|end\_header\_id|>\textbackslash n\textbackslash n don't answer with any format (like markdown), just in natural language.\textbackslash n<|eot\_id|><|start\_header\_id|> user <|end\_header\_id|>\textbackslash n\textbackslash n Classify the following sentence into one of the following emotions: ['anger', 'anticipation', 'disgust', 'fear', 'joy', 'love', 'optimism', 'pessimism', 'sadness', 'surprise', 'trust']. The sentence is: \{\textcolor{red}{sentence}\}. The class of this sentence is? <|eot\_id|><|start\_header\_id|>assistant<|end\_header\_id|>\textbackslash n\textbackslash n The sentence you provided is classified as \textbackslash n['}

\end{tcolorbox}

\begin{tcolorbox}[colback=white, colframe=black, coltitle=white, colbacktitle=black,
    title={System Prompt for LLM-as-Judge}, boxrule=0.5mm, sharp corners]

    \noindent \texttt{\textbf{System:} Your task is to determine if the provided answer is true or false based solely on the ground truth answers given to you in the format \{['answer 1', 'answer 2', \dots]\}. DO NOT rely on your memory; only use the information provided after this instruction. Respond with \{1\} if the predicted answer is correct, which means semantically consistent with any of the ground truth answers, otherwise respond with \{0\}. Respond with just \{0\} or \{1\}, and DO NOT include anything else in your response. This is the only instruction you need to follow.}

    \vspace{10pt}

    \noindent \texttt{\textbf{User:} Input: Who is elected as the vice president of India in 2017?} \\
    \texttt{\textbf{Ground Truth:} \{['Venkaiah Naidu', 'Muppavarapu Venkaiah Naidu']\}} \\
    \texttt{\textbf{Provided Answer:} M. Venkaiah Naidu} \\
    \texttt{\textbf{Assistant:} 1}

    \vspace{10pt}

    \noindent \texttt{\textbf{User:} Input: Who sings 'You are a magnet and I am steel'?} \\
    \texttt{\textbf{Ground Truth:} \{['Walter Egan']\}} \\
    \texttt{\textbf{Provided Answer:} The song ‘You Are a Magnet and I Am Steel’ is performed by the band The 1975.} \\
    \texttt{\textbf{Assistant:} 0}

    \vspace{10pt}

    \noindent \texttt{\textbf{User:} \{\textcolor{red}{prompt}\}}

\end{tcolorbox}
\subsection{Details for Comparison Methods}
For sampling-based uncertainty estimation methods, including: LN-Entropy (LN-E, Semantic Entropy (SE), D-Semantic Entropy (DSE), Lexical Similarity (LeS).We follow the default setting in the original paper and sample $10$ generations with a temperature of $0.5$ to estimate the uncertainty scores. Specifically, for LeS, We use the Rouge-L as the similarity metric, and for SE and DSE, we follow \cite{farquhar2024detecting}, using Deberta-Large-MNLI as our entailment model to get the semantic clusters. For other methods, including Probability and Entropy, we generate the most likely answers using greedy decoding, save the logit values for each token, and further compute their respective uncertainty scores.

\section{Future Work}

\textbf{Logits-induced Sentence Uncertainty.}\quad LogTokU now can capture token uncertainty, but the relationship between token reliability and sentence reliability is still unclear. For example, if there are two sentences where one sentence has a certain degree of uncertainty for each word and the other sentence has only one word with extremely high uncertainty, while the remaining words have very low uncertainty, which sentence is more reliable? The average strategy in this paper may not be the best approach and there may also be better methods to measure the relationship between tokens and sentences.

\textbf{More Modeling Techniques.}\quad Here, we model LogTokU according to logits and the Dirichlet distribution. Other strategies are also optional, such as using energy-based modeling to analyze logits or measuring EU and AU by observing network characteristics, such as the performance of attention.

\textbf{More Applications for Utilizing LogTokU.}\quad In this paper, we present two usage cases and highlight the potential to explore more usage scenarios. For example, we can fine-tune the model based on EU to inject missing knowledge, or improve the model's performance by using the Retrieval Augmented Generation (RAG) strategy to incorporate relevant knowledge as context.

\section{Limitations}

LogTokU demonstrates great potential and may even open up a new space for exploration. However, we must acknowledge that LogTokU has some unavoidable limitations:

\textbf{(1) It cannot estimate the uncertainty of black-box models.}\quad Compared to uncertainty estimation methods that rely on activation states and attention layers, LogTokU requires only logits and can be considered a gray-box uncertainty estimation approach. However, we note that most commercial models do not provide logits as output. Therefore, how to evaluate the uncertainty of black-box LLMs remains an issue for future research.

\textbf{(2) It cannot estimate the uncertainty of the distilled models.}\quad  Distilled models are optimized by using the output probability distribution of a larger model as the learning target, and they often rescale logits at each layer. As a result, small models trained through distillation lose the strength of evidence in their logits. In this case, since logits do not carry evidence strength information, LogTokU cannot be used for uncertainty modeling.

\section{Theoretical Analyze of Why Logits can Capture Epistemic Uncertainty}

\begin{tcolorbox}[colback=gray!10,
                  colframe=black,
                  width=\linewidth,
                  arc=1mm, auto outer arc,
                  boxrule=0.5pt,
                 ]
\begin{theorem}
For any LLM \(\mathcal{M}\) trained with the cross-entropy loss \(L_{\text{CE}}\) using gradient descent optimization (i.e., \(\nabla_\mathcal{M} L_{\text{CE}}\)), the total evidence \( \sum_{\tau^i \in \mathcal{T}} z_{\tau^i} \) will strictly accumulate (i.e., \( \Delta\sum_{\tau^i \in \mathcal{T}} z_{\tau^i}>0 \)), thus the epistemic uncertainty defined in Eq.~\ref{eq:eu} will strictly decrease (i.e., \( \Delta EU \leq 0 \)).
\label{thm:main}
\end{theorem}
\end{tcolorbox}

\textit{Proof.}
Let \( z_{\tau^1}, z_{\tau^2}, \dots, z_{\tau^{|\bm{Y}|}} \) denote the logits corresponding to the classes \( \bm{Y}=\{\tau^1,\tau^2,\cdots,\tau^{|\bm{Y}|}\} \). The softmax function computes the class probabilities \( p_{\tau^1}, p_{\tau^2}, \dots, p_{\tau^{|\bm{Y}|}} \) as follows:

\[
p_{\tau^i} = \frac{e^{z_{\tau^i}}}{\sum_{j=1}^{|\bm{Y}|} e^{z_{\tau^j}}}.
\]

Let \( y \) represent the true label of the input, encoded as a one-hot vector. The cross-entropy loss \( L \) for a single training example is given by:

\[
L_{\text{CE}} = -\sum_{i=1}^{|\bm{Y}|} y_{\tau^i} \log(p_{\tau^i}).
\]

Since \( y \) is one-hot encoded, only the term corresponding to the correct class \( k \) is non-zero. Thus, the loss simplifies to:

\[
L_{\text{CE}} = -\log(p_{\tau^k}).
\]

Substituting \( p_{\tau^k} \) with its expression from the softmax function yields:

\[
L_{\text{CE}} = -\log\left(\frac{e^{z_{\tau^k}}}{\sum_{j=1}^{|\bm{Y}|} e^{z_{\tau^j}}}\right).
\]

To analyze the effect of gradient updates, we compute the gradient of the loss \( L \) with respect to the logits \( z_i \). We consider two cases: (1) \( \tau^i = \tau^k \) (the correct class) and (2) \( \tau^i \neq \tau^k \) (incorrect classes).

\textit{1. Gradient for the correct class (\( \tau^i = \tau^k \)):}

\[
\frac{\partial L}{\partial z_{\tau^k}} = -\left(1 - p_{\tau^k}\right).
\]

\textit{2. Gradient for incorrect classes (\( \tau^i \neq \tau^k \)):}

\[
\frac{\partial L}{\partial z_{\tau^i}} = p_{\tau^i}.
\]

These gradients describe how the loss changes with respect to the logits. Using gradient descent with a learning rate \( \eta \), the update rules for the logits are as follows:

\textit{1. Update for the correct class (\( {\tau^i} = {\tau^k} \)):}

\[
z_{\tau^k} := z_{\tau^k} + \eta (1 - p_{\tau^k}).
\]

\textit{2. Update for incorrect classes (\( {\tau^i} \neq {\tau^k} \)):}

\[
z_{\tau^i} := z_{\tau^i} - \eta p_{\tau^i}.
\]

These updates ensure that the logit for the correct class increases, while the logits for incorrect classes decrease. 

Next, we analyze the change in the total evidence, defined as the sum of the $k$-largest logits (defined in Eq.~\ref{eq:evidence}). Let \( \mathcal{T} \) represent the set of classes corresponding to the top predicted classes (\(\tau^k \in \mathcal{T}\)). For any optimization step, the change in the sum of logits is given by:

\[
\Delta\sum_{\tau^i \in \mathcal{T}} z_{\tau^i} := \Delta z_{\tau^k} + \sum_{\tau^i \in \mathcal{T}, \tau^i \neq \tau^k} \Delta z_{\tau^i}.
\]

Substituting the update rules, we obtain:

\begin{equation}
\begin{aligned}
\Delta\sum_{\tau^i \in \mathcal{T}} z_{\tau^i} & = \eta(1 - p_{\tau^k}) + \sum_{\tau^i \in \textcolor{brown}{\mathcal{T}}, \tau^i \neq \tau^k} (-\eta p_{\tau^i}) \\ & \textcolor{red}{\bm{\geq}} ~\eta(1 - p_{\tau^k}) + \sum_{\tau^i \in \textcolor{brown}{\bm{Y}}, \tau^i \neq \tau^k} (-\eta p_{\tau^i}) \\ &= \eta(1 - p_{\tau^k}) - \eta(1 - p_{\tau^k})
\\&= 0.    
\end{aligned}    
\end{equation}

Since \( \mathcal{T} \) is a subset of the full set of classes \( \bm{Y} \), and \( |\mathcal{T}| \leq |\bm{Y}| \), the inequality \( \Delta\sum_{\tau^i \in \mathcal{T}} z_{\tau^i} \geq 0 \) holds. This implies that the total evidence increases, leading to a decrease in epistemic uncertainty, i.e., \( \Delta EU \leq 0 \).

\end{document}